\documentclass[conference]{IEEEtran}
\IEEEoverridecommandlockouts
\usepackage{cite}
\usepackage{amsmath,amssymb,amsfonts}
\usepackage{algorithmic}
\usepackage{graphicx}
\usepackage{textcomp}
\usepackage{xcolor}
\usepackage{multirow}
\def\BibTeX{{\rm B\kern-.05em{\sc i\kern-.025em b}\kern-.08em
    T\kern-.1667em\lower.7ex\hbox{E}\kern-.125emX}}
\begin{document}

\title{Camouflaged Object Detection \\with Feature Grafting and Distractor Aware
\thanks{*Corresponding author. This work is supported in part by the National Natural Science Foundation of China (Grant No. 41927805).}
}

\author{\IEEEauthorblockN{Yuxuan Song}
\IEEEauthorblockA{\textit{College of Computer} \\
\textit{Science and Technology} \\
\textit{Ocean University of China}\\
Qingdao, China \\
syxvision@stu.ouc.edu.cn}
\and
\IEEEauthorblockN{Xinyue Li}
\IEEEauthorblockA{\textit{College of Computer} \\
\textit{Science and Technology} \\
\textit{Ocean University of China}\\
Qingdao, China \\
lixinyue3550@stu.ouc.edu.cn}
\and
\IEEEauthorblockN{Lin Qi*}
\IEEEauthorblockA{\textit{College of Computer} \\
\textit{Science and Technology} \\
\textit{Ocean University of China}\\
Qingdao, China \\
qilin@ouc.edu.cn}
}

\maketitle

\begin{abstract}
The task of Camouflaged Object Detection (COD) aims to accurately segment camouflaged objects that integrated into the environment, which is more challenging than ordinary detection as the texture between the target and background is visually indistinguishable. In this paper, we proposed a novel Feature Grafting and Distractor Aware network (FDNet) to handle the COD task. Specifically, we use CNN and Transformer to encode multi-scale images in parallel. In order to better explore the advantages of the two encoders, we design a cross-attention-based Feature Grafting Module to graft features extracted from Transformer branch into CNN branch, after which the features are aggregated  in the Feature Fusion Module. A Distractor Aware Module is designed to explicitly model the two possible distractor in the COD task to refine the coarse camouflage map. We also proposed the largest artificial camouflaged object dataset which contains 2000 images with annotations, named ACOD2K. We conducted extensive experiments on four widely used benchmark datasets and the ACOD2K dataset. The results show that our method significantly outperforms other state-of-the-art methods. The code and the ACOD2K will be available at https://github.com/syxvision/FDNet.
\end{abstract}


\section{Introduction}
\label{sec:intro}

Camouflage refers to creatures use the similarity of color, texture, etc. to hide themselves in the background without being discovered by predators. Inspired by the natural camouflage of animals such as chameleon, artificial camouflage was created to deceive human's visual inspection. The computer vision task of Camouflaged Object Detection (COD) aims to accurately segment concealed objects from the background environment, which has recently attracted interests of researchers and facilitated many applications in different fields. However, due to its inherent nature, locating and segmenting of camouflaged objects is much more difficult than ordinary object detection, which makes the COD task extremely challenging.

\begin{figure}[ht]
    \centering
    \includegraphics[width=\columnwidth]{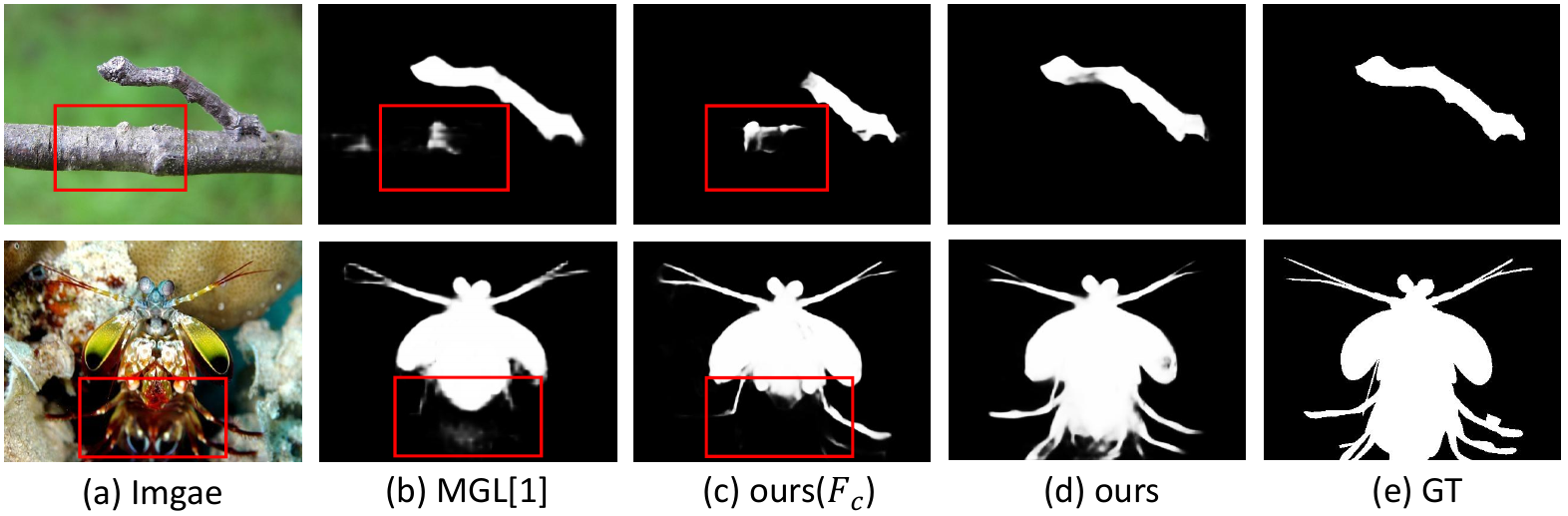}
    \caption{We show two different types of distractors. MGL\cite{mgl}(b) and coarse prediction of our method (c) wrongly detect non-COD regions as COD regions (fp distractor), and also wrongly identify COD regions as non-COD regions (fn distractor). By refining the coarse prediction map with a Distractor Aware Module, our method(d) removes distractors well.}
    \label{fig:distractor}   
\end{figure}

Recently, many deep learning based methods have been proposed to solve the COD task and have achieved impressive progress. SegMaR~\cite{segmar} introduces a Magnification Module to iteratively upsample images to segment camouflaged objects with complex structures. ZoomNet~\cite{ZoomNet} showed that multi-scale information is very effective for resolving the appearance and shape variation of objects at different scales. This model uses a shared encoder to encode images of three scales. However, shared encoders cannot take full advantage of multi-scale images and may cause error propagation. Therefore, we proposed use two different encoders in parallel, and designed a Feature Grafting Module for better feature transfer.

Existing COD methods only consider the background as distractor, such as SINetv2~\cite{sinet2} which uses reverse attention to erase the foreground and use the background to mine potential camouflage areas. However, in the COD task, due to the similarity between the object and the surrounding environment, there are two different types of distractors as shown in Figure \ref{fig:distractor}: 1) in the first row, the stem of the branch is misclassified as camouflaged object since its texture is very similar to the target. 2) in the second row, the lower half of the animal's body is blended with the black background, and the network misses it. This observation inspired us that explicitly modeling semantic features of these two types of distractors with supervision can improve detection performance.

In this paper, we propose a Feature Grafting and Distractor Aware network (FDNet) for camouflaged object detection. We employ Transformer and CNN to exploit information on different scales, where Transformer models long-term dependence for rich context information and CNN mines local details for edge information. To aggregate the features from these two encoders, we developped a Feature Grafting Module based on cross-attention, which fuses features in a bottom-up manner to produce a coarse prediction map. A Distractor Aware Module was designed to guide the learning by modeling the two types of distractor and exploring potential camouflage regions under the supervision of groundtruth. Benefited from the designed modules, our proposed network can better recognize distractors and achieve better detection performance.

\begin{figure}[h]
    \centering
    \includegraphics[width=\columnwidth]{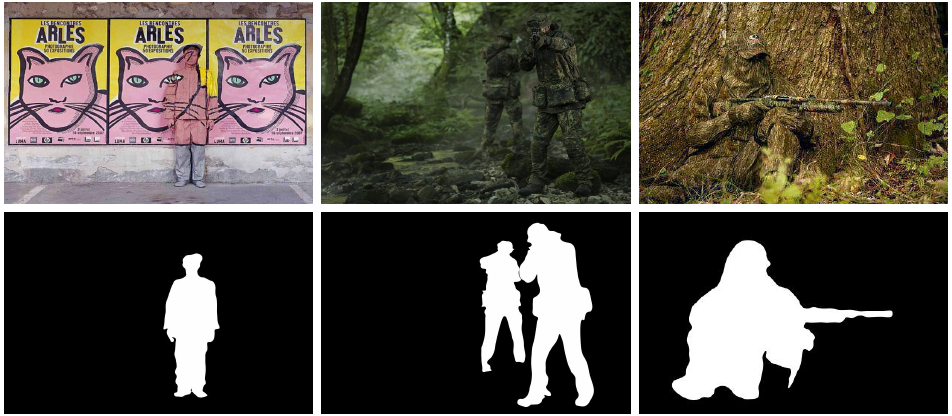}
    \caption{Examples of the artificial camouflages in the proposed ACOD2K dataset.}
    \label{fig:example}   
\end{figure}

In addition, we contribute to the COD community with a new COD dataset under the fact that most existing COD datasets consists of natural camouflaged animals, whereas only a small portion are camouflage created by human. To address this limitation, we collected and annotated 2000 images of artificial camouflages from the Internet, constituting the current largest artificial camouflage dataset, named ACOD2K. Figure \ref{fig:example} shows some exmaple images of this dataset. We compared our proposed model with other state-of-the-art models on public datasets and this new dataset. 

\noindent \textbf{Our contributions.}
1) Camouflaged objects can be segmented more accurately by our proposed FDNet which featured by the multi-scale feature extractor and the explicitly modeling of distractors. 2) The parallel encoding and the Feature Grafting Module are able to extract and fuse multi-scale features, which are utilized by the Distractor Aware Module to incorporate two different types of distracting semantic cues for target segmentation. 3) A large artificial camouflage dataset, ACOD2K, was proposed and tested to compare the performance of our proposed model and other existing models.

\begin{figure*}[ht]
    \centering
    \includegraphics[width=0.9\textwidth]{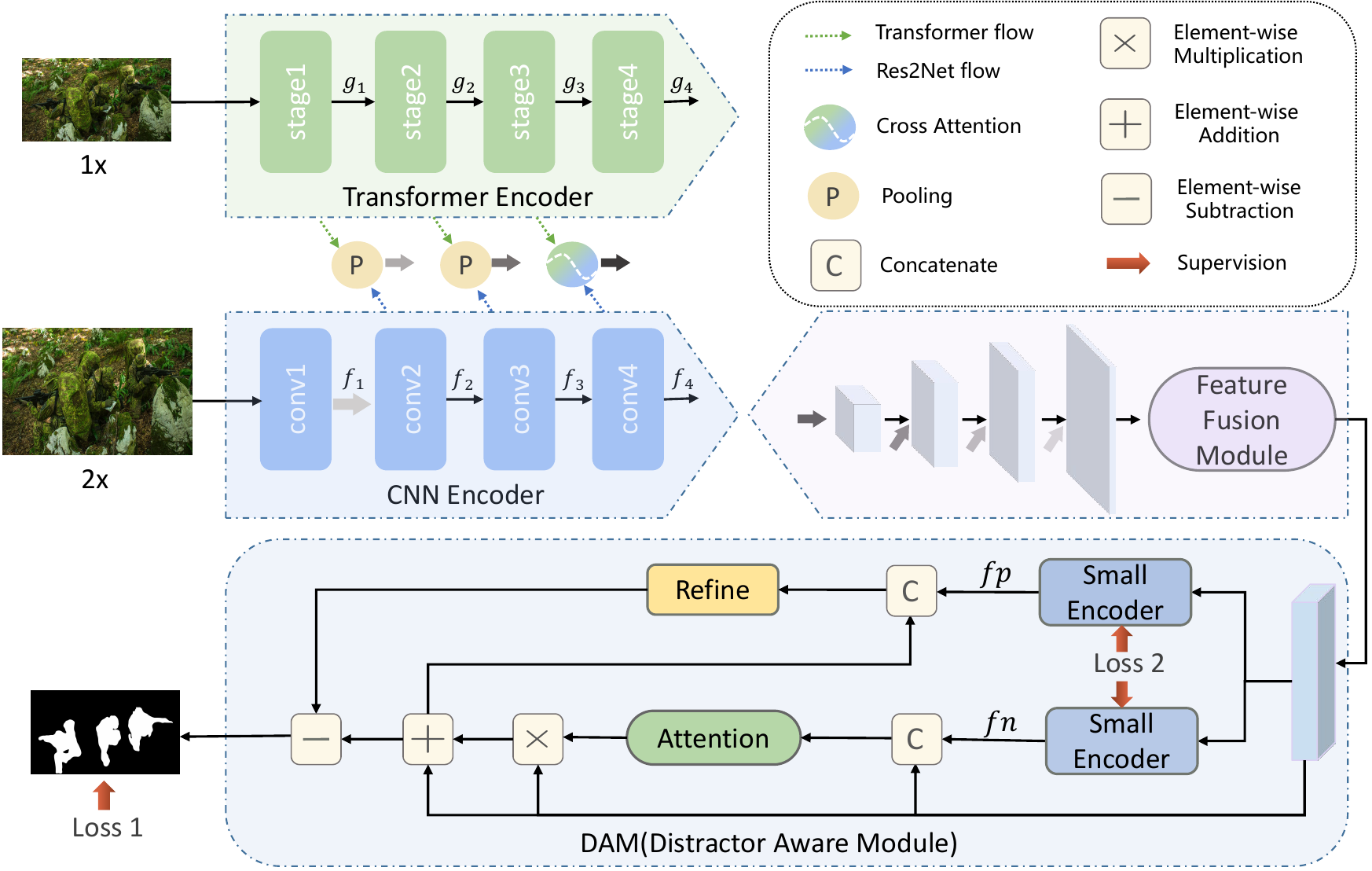}
    \caption{Overview of our proposed Feature Grafting and Distractor Aware Network(FDNet) and its three modules: (a) Feature Grafting Module. (b) Feature Fusion Module. (c) Distractor Aware Module.}
    \label{fig:Net}   
\end{figure*}
\section{Related Work}

The release of large-scale camouflage datasets (such as COD10K~\cite{sinet}) has triggered the invention of many deep learning-based methods, which have shown impressive results for the COD task. A majority of the recent work are inspired by how human observers visually search camouflaged targets, as SINet~\cite{sinet}, ZoomNet~\cite{ZoomNet} and SegMaR~\cite{segmar}. SINet was designed to have two stages for searching and recognition respectively. ZoomNet~\cite{ZoomNet} and the recently proposed SegMaR~\cite{segmar} enlarge the image in potential target regions to further mine distinguishing clues in a coarse-to-fine manner. Other work proposed to use auxiliary cues to improve performance, such as making better use of boundary clues~\cite{bgnet} and frequency-domain perceptual cues~\cite{frequency}. The joint task learning was also found to be useful when SOD(Salient Object Detection) and COD are simultaneously considered to boost each other's performance~\cite{ujsc}.

Unlike CNN, Transformer has a global receptive fields, which can capture richer contextual information. Its success in the natural language processing has been observed by computer vision tasks. UGTR~\cite{ugtr} uses Bayesian and Transformer to infer areas of uncertainty. To take the advantage of both architecture, we employ CNN and Transformer together to enhance the performance of the model.

\section{Our Method}
\subsection{ACOD2K dataset}
\label{sec:dataset}
Camouflage images can be categorized as natural or artificial. Natural camouflage refers to the ability of animals to blend into their surroundings through changes in their physiological characteristics, making them difficult to detect by predators. Artificial camouflage refers to camouflage designed using human reasoning through methods such as painting and camouflage uniforms, with a specific aim to target human visual perception characteristics in order to more effectively deceive the human visual system. It has great practical value for tasks such as disaster-assisted search and rescue operations. Leveraging this advantage, we have constructed ACOD2K, the largest artificial camouflage dataset.\par
It's worth noting that current camouflaged object detection methods are exclusively trained on natural camouflaged images. This is because existing datasets mainly feature natural camouflaged animals, making it difficult to train models that can accurately detect artificial camouflage. For instance, the two most commonly used training datasets in COD tasks, CAMO and COD10K, have an imbalanced distribution of natural and artificial camouflage images. Of the 2,500 images in CAMO, less than 10$\%$ are artificial camouflage images. Similarly, COD10K, a large-scale dataset with 10,000 images covering multiple camouflaged objects in natural scenes divided into 5 super classes, lacks artificial camouflage images. This highlights the need for datasets like ACOD2K, which has a significant number of artificial camouflage images, to enable the development of more robust camouflaged object detection methods.\par
ACOD2K are consisted by 2000 images, where 1500 images are with camouflaged objects, 400 images are with non-camouflaged objects, and 100 are background images. Most of the images are collected from the Internet ($80\%$), searched using the keywords such as ``military camouflage'', ``body painting'', ``Ghillie suit'', and the rest are from public COD and SOD dataset. Figure \ref{fig:example} shows some examples of ACOD2K, from which it can be seen that artificial camouflages are intentionally made by humans using materials and colors to conceal the whole target body in the background. High-quality and fine-grained pixel-level matting annotations were carried out for each image. In order to guarantee the quality, an additional researcher further verified all annotations. 

\subsection{Overall Architecture}
\label{sec:architecture}
The overall structure of our proposed FDNet is shown in Figure \ref{fig:Net}. It is divided into two stages, the first stage generates a coarse feature map, and the second stage refines the feature map based on the Distractor Aware Module. FDNet uses multi-scale images as input. Unlike ZoomNet which uses shared encoders, we instead used the PVT~\cite{pvt} for the main scale and used the Res2Net50~\cite{res2net} for the sub-scale, which constitue a parallel encoder. We designed a Feature Grafting Module based on cross-attention to aggregate features of these two scales, which not only extracts valuable semantic clues, but also fully suppresses redundant information and background noise. Then the multi-scale features are sent to the Feature Fusion Module for decoding, it achieved more efficient transmission of encoded information through bottom-up dense connections. Finally, Send it into the dual-branch Distractor Aware Module to refine the feature map, and use ground truth for supervision.

\begin{figure}[h]
    \centering
    \includegraphics[width=\columnwidth]{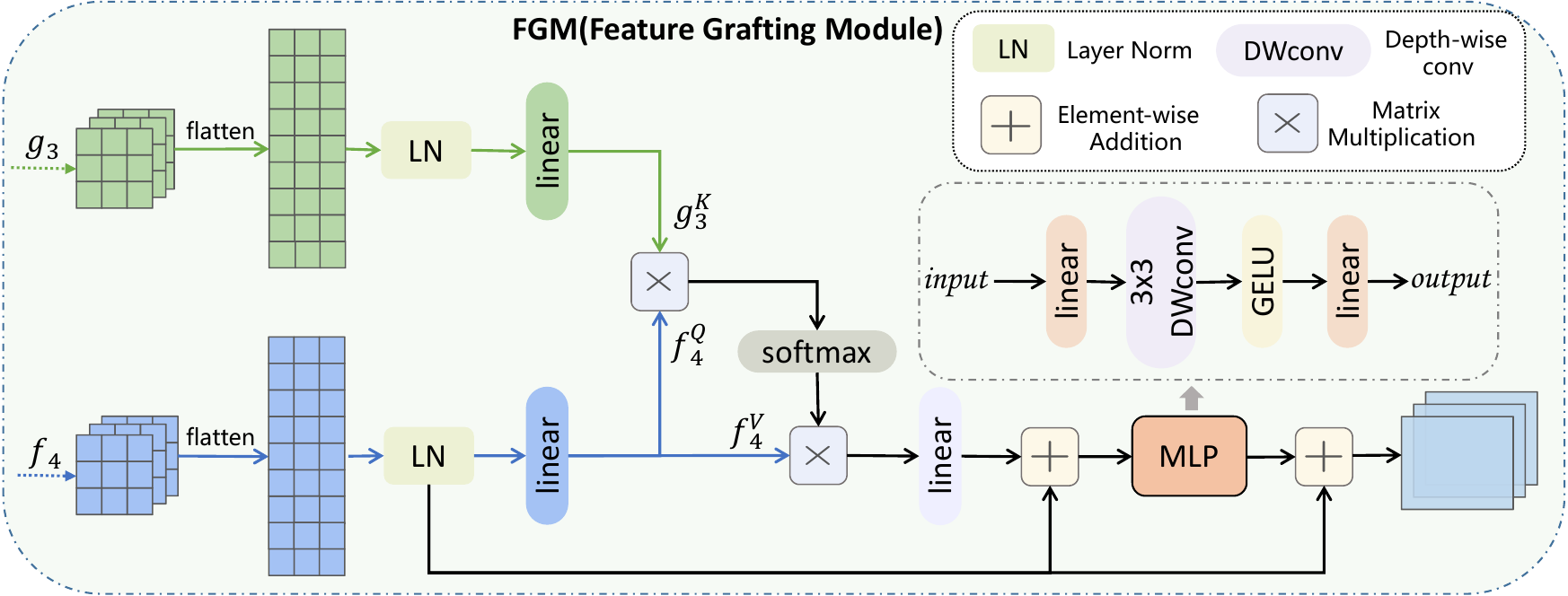}
    \caption{cross-attention-based Feature Grafting Module}
    \label{fig:cross_attention}   
\end{figure}

\subsection{Feature Grafting Module}
\label{sec:grafting}
For the main scale image, we use PVT as the backbone to extract feature maps of 4 stages, which can be denoted as ${g_i;i=1,2,3,4}$. Since the features with too small resolution will lose most of the information, we did not use $g_4$. For the sub-scale image, we use Res2Net50 as the backbone to extract a set of feature maps, which can be denoted as ${f_i;i=1,2,3,4}$.\par
We choose to graft feature on feature groups with the same feature resolution. Since the resolution of the sub-scale is twice that of the main scale, the resolution of ${g_i,f_{i+1};i=1,2,3}$ is same. For the first two groups, we use pooling for feature grafting to maintain and highlight useful information. In neural networks, deeper features have richer semantic clues. For $g_3$ extracted using Transformer, which has rich global context information. For $f_4$ extracted using CNN, which has edge detail information complementary to global information. We believe that using simple fusion methods such as pooling, concatenation, or addition is not effective enough for mutual learning between these two features, and cannot well suppress background noise from CNN. Therefore, we use cross-attention to incorporate the global semantic cue learned from the main scale into each pixel of the sub-scale. The detail is shown in Figure \ref{fig:cross_attention}.\par

\begin{equation}
F_4 = Softmax(\frac{f_4^Q \cdot {g_3^K}^T}{\sqrt{k}})\cdot f_4^V\
\label{eq:F4}
\end{equation}

\begin{equation}
f_4^Q,f_4^V=\theta(f_4) \qquad g_3^K=\phi(g_3)
\end{equation}

$\theta()$ uses flatten and permute operations to transform $f_4\in R^{C \times H \times W}$ into $f_4^{'}\in R^{HW \times C}$. Same as self attention, we have passed it through Layer Normalization and linear transformation to get $f_4^Q, f_4^V$, the process of $g_3$ getting $g_3^K$ through $\phi()$ is same as $\theta$.

\subsection{Feature Fusion Module}
\label{sec:fusion}
\begin{figure}[h]
    \centering
    \includegraphics[width=\columnwidth]{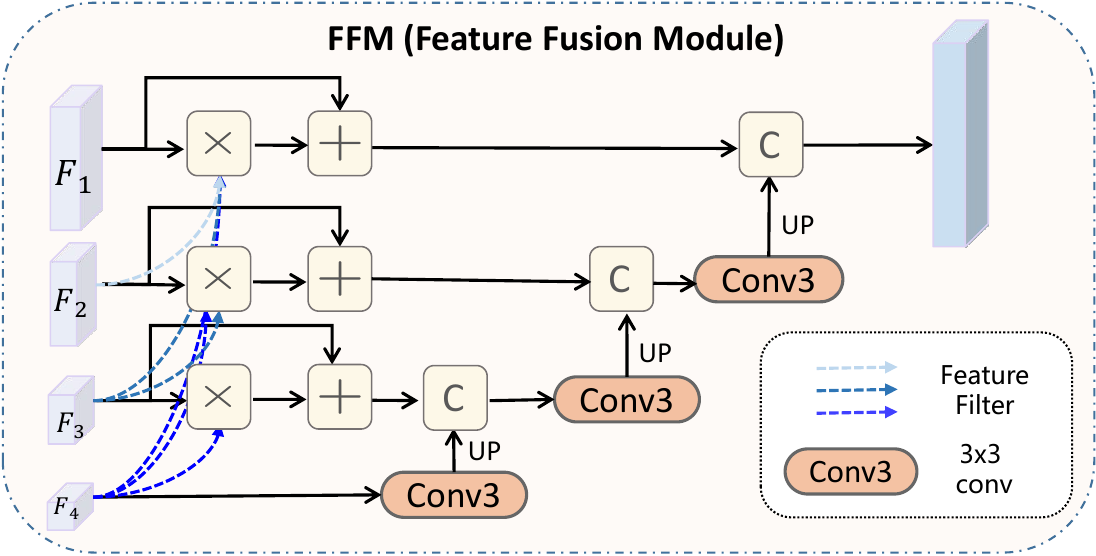}
    \caption{Feature Fusion Module}
    \label{fig:feature_fusion}   
\end{figure}

Unlike the previous method that directly performs convolution after channel concat on the adjacent feature layer to output the prediction map, we fuse deeper features as a semantic filter. We first element-wise multiply it with the current layer features to suppress background interference that may cause abnormality, and then preserve the original information by residual addition. The details are shown in Figure \ref{fig:feature_fusion}.

The features by the Feature Grafting Module are denoted as $F_i;i=1,2,3,4$. Since $F_4$ is the last layer of features, we directly perform 3x3 convolution on $F_4$ to form $\hat{F_4}$, For $F_3$, we perform filtering on F4 to form $F_3^{filter}$. Correspondingly, $F_2^{filter}$ and $F_1^{filter}$ are shown in the following formula. We take the top-level feature $\hat{F_1}$ as the final result of the Feature Fusion Module, and the coarse prediction is $F_c$.\par

\begin{equation}
\hat{F_4} = Conv3(F_4)
\end{equation}
\begin{equation}
F_3^{filter} = Conv3(Conv1(F_4\uparrow_2)
\end{equation}
\begin{equation}
\hat{F_3} = Conv3([F_3^{filter} * F_3+F_3;\hat{F_4}])
\end{equation}
\begin{equation}
F_2^{filter} = Conv3(Conv1([F_4\uparrow_4;F_3\uparrow_2]))
\end{equation}
\begin{equation}
\hat{F_2} = Conv3([F_2^{filter} * F_2+F_2;\hat{F_3}])
\end{equation}
\begin{equation}
F_1^{filter} = Conv3(Conv1([F_4\uparrow_8;F_3\uparrow_4;F_2\uparrow_2]))
\end{equation}
\begin{equation}
\hat{F_1} = Conv3([F_1^{filter} * F_1+F_1;\hat{F_2}])
\end{equation}
\begin{equation}
F_c=Conv3(\hat{F_1})
\end{equation}

Conv3, Conv1 represents 3x3, 1x1 convolution respectively, $\uparrow$ refers to upsample, [;] means channel concatenation, and * represents element-wise multiplication.

\subsection{Distractor Aware Module}
\label{sec:distractor} 
We believe that there are two types of distractors present in the coarse prediction map generated in the first stage, namely: (i) objects that are camouflaged but not detected, referred to as false negatives, $\xi_{fn}$, and (ii) objects that are not camouflaged but are misdetected, referred to as false positives, $\xi_{fp}$. To address this, we propose a dual-branch Distractor Aware Module that explicitly models the potential interference and aims to improve the accuracy of the segmentation results.\par
As illustrated in the lower part of Figure \ref{fig:Net}, we first use $\hat{F_1} \in R^{64 \times H \times W}$ to extract $\xi_{fn}$ features through a lightweight encoder, the encoder is designed as two 3x3 convolutions, following BN and Relu. In order to make better use of $\xi_{fn}$, We generated the predicted map of $\xi_{fn}$. During training, the ground truth of $\xi_{fn}$ is approximated by the difference between the ground truth of the segmentation map and the coarse predicted map $F_c$. Then we concate $\xi_{fn}$ with $\hat{F_1}$ and send it into the attention mechanism to generate augmented weights $\xi_{fn}^a$. The attention mechanism aims to enhance the features of possible $\xi_{fn}$ regions. we perform element-wise multiplication for $\xi_{fn}^a$ and original feature $\hat{F_1}$, and then perform residual connection to generate the enhanced feature $F_{fn}$. Now, the network can better segment those regions that are ignored as background.\par

\begin{equation}
\xi_{fn} = Small\ Encoder(\hat{F_1})
\end{equation}
\begin{equation}
{fn}_{GT} = GT - \varphi(F_c)
\end{equation}

Similarly, we use the same encoder to extract $\xi_{fp}$ features and the predicted map. The ground truth of $\xi_{fp}$ is approximated by the difference between the coarse predicted map $F_c$ and the ground truth of the segmentation map. we concate $F_{fn}$ with $\xi_{fp}$ on channel dimension, then send it into the refine unit consisting of two 3x3 convolutional layers to capture richer context information, so as to better distinguish the misdetected areas. Finally, it is subtracted from $F_{fn}$ to obtain the prediction feature that suppresses $\xi_{fp}$ distractor. After 3x3 convolution, we obtain the final prediction map $F_{p}$. $\varphi()$ represents binarization operation.
\begin{equation}
\xi_{fp} = Small\ Encoder(\hat{F_1})
\end{equation}
\begin{equation}
{fp}_{GT} = \varphi(F_c) - GT
\end{equation}

\begin{table*}[h]
\caption{Quantitative comparison with 10 methods on four datasets.The best results are highlighted in \textbf{Bold}.}
\resizebox{\textwidth}{!}{
\begin{tabular}{c|c|cccc|cccc|cccc|cccc}
\hline
\multirow{2}{*}{Method} & 
\multirow{2}{*}{Year}  & 
\multicolumn{4}{c|}{CAMO-Test} & 
\multicolumn{4}{c|}{CHAMELEON} &
\multicolumn{4}{c|}{COD10K-Test} & 
\multicolumn{4}{c}{NC4K} \\ \cline{3-18}
\multicolumn{1}{c|}{} & \multicolumn{1}{c|}{} & $S_\alpha\uparrow$ & $F_\beta^w\uparrow$ & $E_\phi\uparrow$ & $\mathcal{M}\downarrow$ &
 $S_\alpha\uparrow$ & $F_\beta^w\uparrow$ & $E_\phi\uparrow$ & $\mathcal{M}\downarrow$ &
 $S_\alpha\uparrow$ & $F_\beta^w\uparrow$ & $E_\phi\uparrow$ & $\mathcal{M}\downarrow$ &
 $S_\alpha\uparrow$ & $F_\beta^w\uparrow$ & $E_\phi\uparrow$ & $\mathcal{M}\downarrow$  \\ \hline
F$^3$Net\cite{f3net} & 2020 & 0.711 & 0.564 & 0.741 & 0.109 & 0.848 & 0.744 & 0.894 & 0.047 & 0.739 & 0.544 & 0.795 & 0.051 & 0.780 & 0.656 & 0.824 & 0.070\\
PraNet~\cite{pranet} & 2020 & 0.769 & 0.663 & 0.837 & 0.094 & 0.860 & 0.763 & 0.935 & 0.044 & 0.789 & 0.629 & 0.879 & 0.045 & 0.822 & 0.724 & 0.888 & 0.059\\
CSNet~\cite{csnet} & 2020 & 0.771 & 0.642 & 0.795 & 0.092 & 0.856 & 0.718 & 0.869 & 0.047 & 0.778 & 0.569 & 0.810 & 0.047 & 0.750 & 0.603 & 0.773 & 0.088\\
SINet~\cite{sinet} & 2020 & 0.751 & 0.606 & 0.771 & 0.100 & 0.869 & 0.740 & 0.891 & 0.044 & 0.771 & 0.551 & 0.806 & 0.051 & 0.808 & 0.723 & 0.883 & 0.058\\
PFNet~\cite{pfnet} & 2021 & 0.782 & 0.695 & 0.852 & 0.085 & 0.882 & 0.810 & 0.942 & 0.033 & 0.800 & 0.660 & 0.868 & 0.040 & 0.829 & 0.745 & 0.898 & 0.053\\ 
C$^2$Net~\cite{c2fnet} & 2021 & 0.796 & 0.719 & 0.864 & 0.080 & 0.888 & 0.828 & 0.946 & 0.032 & 0.813 & 0.686 & 0.900 & 0.036 & 0.838 & 0.762 & 0.904 & 0.049\\ 
LSR~\cite{lsr} & 2021 & 0.787 & 0.696 & 0.854 & 0.080 & 0.890 & 0.822 & 0.948 & 0.030 & 0.804 & 0.673 & 0.892 & 0.037 & 0.840 & 0.766 & 0.907 & 0.048\\
UGTR~\cite{ugtr} & 2021 & 0.784 & 0.684 & 0.851 & 0.086 & 0.888 & 0.794 & 0.940 & 0.031 & 0.817 & 0.666 & 0.890 & 0.036 & 0.839 & 0.746 & 0.899 & 0.052\\ 
UJSC~\cite{ujsc} & 2021 & 0.800 & 0.728 & 0.873 & 0.073 & 0.891 & 0.833 & 0.955 & 0.030 & 0.809 & 0.684 & 0.891 & 0.035 & 0.842 & 0.771 & 0.907 & 0.047\\ 
BgNet~\cite{bgnet} & 2022 & 0.832 & 0.762 & 0.884 & \textbf{0.065} & 0.894 & 0.823 & 0.943 & 0.029 & 0.826 &0.703 & 0.898 & 0.034 & 0.855 & 0.784 & 0.907 & 0.045\\
FDNet~(Ours) & 2022 & \textbf{0.836} & \textbf{0.777} & \textbf{0.886} & 0.066 & \textbf{0.909} & \textbf{0.856} & \textbf{0.947} & \textbf{0.025} & \textbf{0.857} & \textbf{0.763} & \textbf{0.918} & \textbf{0.028} & \textbf{0.865} & \textbf{0.803} & \textbf{0.911} & \textbf{0.042}\\ \hline
\end{tabular}}
\label{tab:quantitative results}
\end{table*}
\begin{table}[ht]
\caption{Quantitative comparison with 3 methods on ACOD2K.The best results are highlighted in \textbf{Bold}.}
\tiny
\centering
\resizebox{0.4\textwidth}{!}{
\begin{tabular}{c|cccc}
\hline
\multirow{2}{*}{Method} & \multicolumn{4}{c}{ACOD2K-Test} \\ \cline{2-5} 
&$S_\alpha\uparrow$ & $F_\beta^w\uparrow$ & $E_\phi\uparrow$  & $\mathcal{M}\downarrow$\\ \hline
SINet  & 0.824 & 0.681 &  0.865 & 0.061 \\
PFNet  & 0.851 & 0.786 & 0.911  & 0.048 \\
C$^2$Fet  & 0.858 & 0.799& 0.912 & 0.046\\
ours   & \textbf{0.872} &\textbf{0.823}  & \textbf{0.922}  & \textbf{0.044} \\ \hline
\end{tabular}}
\label{tab:acod2k}
\end{table}
\begin{figure*}[h]
    \centering{
    }
	\includegraphics[width=1\textwidth]{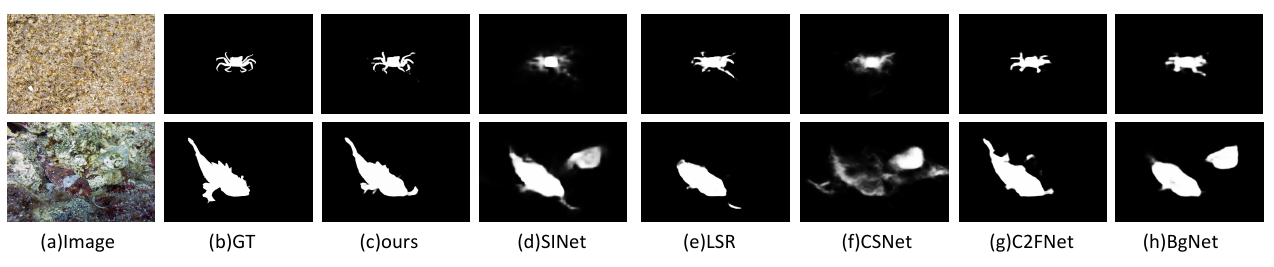}
	\caption{Qualitative comparison with five state-of-art COD methods.}
    \label{fig:results}
\end{figure*}

\subsection{Loss Functions}
\label{sec:loss}
Our network has two types of supervision. For the loss $L_{F_p}$ of the prediction map, same as most COD methods, we use the weighted BCE loss and the weighted IOU loss(Loss1). For the loss $L_{fn}$, $L_{fp}$ of fn and fp, we use the weighted BCE loss(Loss2). The loss function is as follows.\par

\begin{equation}
Loss = L_{F_p}+ \lambda L_{fn} + \beta L_{fp}
\end{equation}
\begin{equation}
Loss1 = L_{BCE}^{\omega}+L_{IOU}^{\omega}
\end{equation}
\begin{equation}
\begin{split}
Loss2 = \sum_i(-[\frac{N_p}{N_p+N_n}(y_i)log(p_i)+\\
\frac{N_n}{N_p+N_n}(1-y_i)log(1-p_i)])    
\end{split}
\end{equation}

In the experiment, $\lambda$ and $\beta$ are set to 10. $N_n$ and $N_p$ represent the number of pixels of positive pixels and negative pixels, respectively. 

\section{Experiments}
\subsection{Experiment Setup}
\noindent \textbf{Datasets.}\par
We perform experiments on four COD benchmark datasets and ours ACOD2K. Public datasets include CAMO~\cite{camo}, CHAMELON~\cite{chameleon}, COD10K~\cite{sinet} and NC4K~\cite{lsr}, Like the previous methods, we use 3040 images from COD10K and 1000 images from CAMO for training, and other datasets for testing. For the ACOD2K, we divide it into train set and test set according to the ratio of 8:2.

\noindent \textbf{Evaluation Criteria.}\par
We use four metrics that commonly used in COD tasks to evaluate the model performance: Mean absolute error(MAE)~\cite{mae}, $F_\beta^w$-measure~\cite{fm}, E-measure~\cite{em}, S-measure~\cite{sm}.

\noindent \textbf{Implementation Details.}\par
Our network uses PVT~\cite{pvt} and Res2Net50~\cite{res2net} pretrained on ImageNet as backbone. We use data augmentation strategy of random flips and rotations. During training, in order to balance efficiency and performance, the size of the main scale is set to 288x288. The batchsize is 32. We use SGD with momentum and weight decay initialized to 0.9 and 0.0005 as the optimizer, the learning rate is initialized to 0.05, follows a linear decay strategy, and the maximum training epoch is set to 50. The entire network is performed on NVIDIA GeForce GTX 3090Ti.

\subsection{Comparisons with State-of-the-arts}
To show the effectiveness of our method, we compare with 10 SOTA methods on public datasets. On ours ACOD2K, we compare with 3 COD methods. For fair comparison, the results of these models are either provided by the authors or retrained from open source code.

\noindent \textbf{Quantitative Evaluation.}\par
As shown in the Table \ref{tab:quantitative results}, our method achieves the superior performance on multiple evaluation metrics. Specifically, our method increases $F_{\beta}^{\omega}$ by 1.5$\%$, 3.3$\%$, 6$\%$, 1.9$\%$ over the second-best method on all four datasets. Table \ref{tab:acod2k} shows the FDNet outperforms the second-best method on the four metrics by increasing 1.4$\%$, 2.4$\%$, 1$\%$,0.4$\%$ on the ACOD2K.

\noindent \textbf{Qualitative Evaluation.}\par
We further show the qualitative comparison of FDNet with other methods, presented in the form of visualization maps. As shown in Figure \ref{fig:results}, our method not only recognizes them well, but also segments fine edges. In addition, in the second row, our method also works well with the presence of distractor in the image.

\begin{table}[h]
\caption{Comparison of ablation experiment details. FFM: Feature Fusion Module. FGM: Feature Grafting Module. DAM: Distractor Aware Module.}
\tiny
\centering
\resizebox{\columnwidth}{!}{
\begin{tabular}{c|ccccc}
\hline
Method & image  & encoder & FFM & FGM & DAM  \\ \hline
A      & single & CNN     &     &     &       \\ 
C      & multi  & CNN     &     &     &       \\ 
B      & single & CNN     & \checkmark   &     &     \\
D      & multi  & CNN+PVT & \checkmark   & \checkmark   &     \\ 
E      & multi  & CNN+PVT & \checkmark   & \checkmark   & \checkmark  \\ \hline
\end{tabular}}
\label{tab:ablation1}

\end{table}
\begin{table}[h]
\caption{Ablation studies on two datasets. The best results are highlighted in \textbf{Bold}.}
\centering
\resizebox{\columnwidth}{!}{
\begin{tabular}{c|cccc|cccc}
\hline
\multirow{2}{*}{Method} &
  \multicolumn{4}{c|}{CAMO-Test} &
  \multicolumn{4}{c}{COD10K-Test}  \\ \cline{2-9} 
\multicolumn{1}{c|}{} &
  $S_\alpha\uparrow$ & $F_\beta^w\uparrow$ & $E_\phi\uparrow$  & $\mathcal{M}\downarrow$ & $S_\alpha\uparrow$ &
  $F_\beta^w\uparrow$ & $E_\phi\uparrow$ & $\mathcal{M}\downarrow$  \\ \hline
A & 0.791 & 0.698 & 0.842 & 0.079 & 0.800 & 0.648 & 0.872 & 0.040 \\ 
B & 0.800 & 0.719 & 0.856 & 0.074 & 0.807 & 0.670 & 0.880 & 0.037 \\ 
C & 0.823 & 0.759 & 0.874 & 0.070 & 0.851 & 0.755 & 0.909 & 0.028 \\ 
D & 0.829 & 0.769 & 0.882 & 0.068 & 0.853 & 0.758 & 0.915 & 0.029 \\ 
E & \textbf{0.836} & \textbf{0.777} & \textbf{0.886} & \textbf{0.066} & \textbf{0.857} & \textbf{0.763} & \textbf{0.918} & \textbf{0.028} \\ \hline
\end{tabular}}
\label{tab:ablation2}
\end{table}
\subsection{Ablation Studies}\label{ablation}
As shown in the Table \ref{tab:ablation1}, we conducted five ablation experiments. In A, we removed all key modules, only used single-scale images, and simply perform convolution after channel concatenation to get the final prediction map. In B, we replaced the Feature Fusion Module on the basis of A. In C, we use multi-scale images, but share the encoder, and the features of different scales are fused by pooling. In D, we use CNN and Transformer to encode the images of two scales respectively, and use the Feature Grafting Module to fuse feature. In E, we added Distractor Aware Module based on D. \par
\noindent \textbf{Effectiveness of multi-scale.}\label{sec:ablation_multi_scale}
By fusing features of different scales, we can explore richer semantic representations. From the second and third rows in the table \ref{tab:ablation2}, it can be seen that the performance of C is significantly better than that of B, especially in the COD10K, $S_\alpha$, $F_\beta^w$ , $E_\phi $, $\mathcal{M}$ increased by 4.4$\%$, 8.5$\%$, 2.9$\%$, 0.9$\%$ respectively.\par
\noindent \textbf{Effectiveness of Feature Fusion.}\label{sec:ablation_fusion}
From the first and second rows of the table \ref{tab:ablation2}, B's performance on the four indicators increased by 0.8$\%$, 2.2$\%$, 1.1$\%$, 0.4$\%$ on average, this is due to the positive impact of the Feature Fusion Module's bottom-up dense feature-guided structure.\par

\noindent \textbf{Effectiveness of Feature Grafting.}\label{sec:ablation_grafting}
Compared with C, all indicators of D on the two datasets have different degrees of increase, especially $F_\beta^w$ on the CAMO increased by 1$\%$. This is largely because Feature Grafting Module aggregates the advantages of two different types of encoders well.\par

\noindent \textbf{Effectiveness of Distractor Aware.}\label{sec:ablation_distractor}
E outperforms D on all datasets, and the visual comparison results in Figure \ref{fig:distractor} also clearly verify that the module can mine potential interference areas.

\section{Conclusion}
We propose a novel COD network, FDNet. First, we design the Feature Grafting Module to extract valuable semantic information and suppress background noise. Then, in the Distractor Aware Module, we obtained more accurate prediction map by refining the two types of distractors. Additionally, we also construct a new artificial camouflage dataset, ACOD2K. Experiments on four public datasets and ACOD2K show that our method outperforms other methods significantly both qualitatively and quantitatively. In the future, we will explore more effective supervision methods for two types of distractors.

\bibliographystyle{IEEEtran}
\bibliography{IEEEexample}

\end{document}